\pdfoutput=1
\documentclass[11pt,a4paper]{article}
\usepackage[hyperref]{emnlp2018}
\usepackage{times}
\usepackage{latexsym}
\usepackage{amsmath}
\usepackage{amssymb}
\usepackage{amsfonts}
\usepackage{graphicx}
\usepackage{url}
\usepackage{stfloats}
\usepackage[inline, shortlabels]{enumitem}
\usepackage{multirow}
\usepackage{array}
\usepackage{mathtools}
\newcolumntype{$}{>{\global\let\currentrowstyle\relax}}
\newcolumntype{^}{>{\currentrowstyle}}
\newcommand{\rowstyle}[1]{\gdef\currentrowstyle{#1}%
  #1\ignorespaces
}

\graphicspath{ {./images/} } 

\aclfinalcopy 

\newcommand\m[1]{\mathbf{#1}}

\title{Understanding Convolutional Neural Networks for Text Classification}

\author{Alon Jacovi$^{1,2}$ \\
   \\\And
   Oren Sar Shalom$^{2,3}$ \\
     $^1$ Computer Science Department, Bar Ilan Univesity, Israel \\
  $^2$ IBM Research, Haifa, Israel \\
  $^3$ Intuit, Hod HaSharon, Israel \\
  $^4$ Allen Institute for Artificial Intelligence \\
  {\tt \{alonjacovi,oren.sarshalom,yoav.goldberg\}@gmail.com} \\\\\And
  Yoav Goldberg$^{1,4}$  \\
   \\}

\date{}

\begin{document}
\maketitle
\begin{abstract}
We present an analysis into the inner workings of Convolutional Neural Networks (CNNs) for processing text. CNNs used for computer vision can be interpreted by projecting filters into image space, but for discrete sequence inputs CNNs remain a mystery. We aim to understand the method by which the networks process and classify text. We examine common hypotheses to this problem: that filters, accompanied by global max-pooling, serve as ngram detectors. We show that filters may capture several different semantic classes of ngrams by using different activation patterns, and that global max-pooling induces behavior which separates important ngrams from the rest. Finally, we show practical use cases derived from our findings in the form of model interpretability (explaining a trained model by deriving a concrete identity for each filter, bridging the gap between visualization tools in vision tasks and NLP) and prediction interpretability (explaining predictions). Code implementation is available online at github.com/sayaendo/interpreting-cnn-for-text.
\end{abstract}

\section{Introduction}

Convolutional Neural Networks (CNNs), originally invented for computer vision, have been shown to achieve strong performance on text classification tasks \cite{DBLP:journals/corr/abs-1803-01271,DBLP:conf/acl/KalchbrennerGB14,DBLP:conf/acl/WangXXLZWH15,DBLP:conf/nips/ZhangZL15,DBLP:conf/naacl/Johnson015,DBLP:conf/acl/IyyerMBD15} as well as other traditional Natural Language Processing (NLP) tasks \cite{DBLP:journals/jmlr/CollobertWBKKK11}, even when considering relatively simple one-layer models \cite{DBLP:conf/emnlp/Kim14}.

As with other architectures of neural networks, explaining the learned functionality of
CNNs is still an active research area. The ability to interpret neural models
can be used to increase trust in model predictions, analyze errors or improve
the model \cite{DBLP:conf/kdd/Ribeiro0G16}. The problem of interpretability in
machine learning can be divided into two concrete tasks: Given a trained model,
\textit{model interpretability} aims to supply a structured \textit{explanation}
which captures what the model has learned. Given a trained model and a single
example, \textit{prediction interpretability} aims to explain how the model
arrived at its prediction. These can be further divided into white-box and
black-box techniques. While recent works have begun to supply the means of
interpreting predictions
\cite{DBLP:conf/emnlp/Alvarez-MelisJ17,DBLP:conf/emnlp/LeiBJ16,DBLP:journals/corr/abs-1805-08969}, interpreting neural NLP models remains an under-explored area.

Accompanying their rising popularity, CNNs have seen multiple advances in
interpretability when used for computer vision tasks
\cite{DBLP:conf/eccv/ZeilerF14}. These techniques
unfortunately do not trivially apply to discrete sequences, as they
assume a continuous input space used to represent images. Intuitions about how
CNNs work on an abstract level also may not carry over from image inputs to
text---for example, pooling in CNNs has been used to induce deformation
invariance \cite{lechun98,lechun15}, which is likely different than the role it
has when processing text.

In this work, we examine and attempt to understand how CNNs process text, and
then use this information for the more practical goals of improving model-level
and prediction-level explanations.

We identify and refine current intuitions as to how CNNs work. Specifically,
current common wisdom suggests that CNNs classify text by working through the
following steps \cite{DBLP:journals/jair/Goldberg16}:
\begin{enumerate}[label=\arabic*)]
\setlength\itemsep{0.1em}
\item 1-dimensional convolving filters are used as ngram detectors, each filter
    specializing in a closely-related family of ngrams.
\item Max-pooling over time extracts the relevant ngrams for making a decision.
\item The rest of the network classifies the text based on this information.
\end{enumerate}

We refine items 1 and 2 and show that:
\begin{itemize}
\setlength\itemsep{0.1em}
    \item Max-pooling induces a thresholding behavior, and values below a given
        threshold are ignored when (i.e. irrelevant to) making a prediction. Specifically, we show an experiment for which 40\% of the pooled ngrams on average can be dropped with no loss of performance (Section \ref{thresholds}).
\item Filters are not homogeneous, i.e. a single filter can, and often does,
    detect multiple distinctly different families of ngrams (Section \ref{clustering}).
\item Filters also detect negative items in ngrams---they not only select for a
    family of ngrams but often actively suppress a related family of negated
    ngrams (Section \ref{negative-ngrams}).
\end{itemize}
We also show that the filters are trained to work with naturally-occurring
ngrams, and can be easily misled (made to produce values substantially larger
than their expected range) by selected non-natural ngrams.

These findings can be used for improving model-level and prediction-level
interpretability (Section \ref{use-cases}). Concretely:
\begin{enumerate*}[label=\arabic*)]
\item We improve model interpretability by deriving a useful summary for each
    filter, highlighting the kinds of structures it is sensitive to.
\item We improve prediction interpretability by focusing on informative ngrams
    and taking into account also the negative cues.
\end{enumerate*}


\section{Background: 1D Text Convolutions} \label{ff-refresh}

We focus on the task of text classification. We consider the common architecture in which each word in a document is
represented as an embedding vector, a single convolutional layer with $m$
filters is applied, producing an $m$-dimensional vector for each document ngram.
The vectors are combined using max-pooling followed by a ReLU activation. The
result is then passed to a linear layer for the final classification.


For an $n$-words input text $w_1,...,w_n$ we embed each symbol as $d$
dimensional vector, resulting in word vectors $\m{w}_1,...,\m{w}_n \in R^d$. The
resulting $d \times n$ matrix is then fed into a convolutional layer where we
pass a sliding window over the text. For each $l$-words ngram: $$ \m{u_i} =
[\m{w}_i,...,\m{w}_{i+\ell-1}] \in R^{d\times\ell} \ ; \ \ 0\leq i \leq n-\ell $$

And
for each filter $\m{f_j} \in R^{d \times \ell}$ we calculate $\langle
\m{u_i},\m{f_j} \rangle$.  The convolution results in matrix $ \m{F} \in R^{n \times
m} $. Applying max-pooling across the ngram dimension results in $\m{p} \in
R^{m}$ which is fed into ReLU non-linearity. Finally, a linear fully-connected
layer $ \m{W} \in R^{c \times m} $ produces the distribution over classification classes from which the strongest class is outputted. Formally: \begin{align*}
    & \m{u}_i = [\m{w}_i;...;\m{w}_{i+\ell-1}] \\
& F_{ij} = \langle \m{u}_i, \m{f}_j \rangle \\
& p_j = \operatorname*{ReLU} (\operatorname*{max}_{i} F_{ij}) \\
& \m{o} = \operatorname*{softmax} (\m{W} \m{p}) 
\end{align*}
In practice, we use multiple window sizes $\ell \in L$, $L \subsetneq \mathbb{N}$ by using multiple convolution layers in parallel and concatenating the resulting $\m{p^\ell}$ vectors. We note that the methods in this work are applicable for dilated convolutions as well.


\section{Datasets and Hyperparameters}
For our empirical experiments and results presented in this work we use three
text classification datasets for Sentiment Analysis, which involves
classifying the input text (user reviews in all cases) between positive and
negative. The specific datasets were chosen for their relative variety in size
and domain as well as for the relative simplicity and interpretability of the
binary sentiment analysis task.

The three datasets are: \begin{enumerate*}[label={\alph*)}]
\item \textbf{MR}: sentence polarity dataset v1.0 introduced by \citet{DBLP:journals/corr/abs-cs-0506075}, containing 10k evenly split short (sentences or snippets) movie reviews.
\item \textbf{Elec}: electronic product reviews for sentiment classification introduced by \citet{DBLP:conf/naacl/Johnson015}, assembled from the Amazon review dataset \cite{DBLP:conf/recsys/McAuleyL13,DBLP:conf/sigir/McAuleyTSH15}, containing 200k train and 25k test evenly split reviews.
\item \textbf{Yelp Review Polarity}: introduced by \citet{DBLP:conf/nips/ZhangZL15} from the Yelp Dataset Challenge 2015, containing 560k train and 38k test evenly split business reviews.
\end{enumerate*}

For word embeddings, we use the pre-trained GloVe \textit{Wikipedia
2014---Gigaword 5} embeddings \cite{pennington2014glove}, which we fine-tune
with the model.

We use embedding dimension of 50, filter sizes of $\ell \in \{2,3,4\}$ words, and $m \in \{10,50\}$
filters. Models are implemented in PyTorch and trained with the Adam optimizer. 


\section{Identifying Important Features} \label{thresholds}



Current common wisdom posits that filters serve as ngram detectors: each filter
searches for a specific class of ngrams, which it marks by assigning them high
scores. These highest-scoring detected ngrams survive the max-pooling
operation. The final decision is then based on the set of ngrams in the
max-pooled vector (represented by
the set of corresponding filters). Intuitively, ngrams which any filter scores
highly (relative to how it scores other ngrams) are ngrams which are highly relevant for the classification of the text.

In this section we refine this view by attempting to answer the questions:
what information about ngrams is captured in the max-pooled vector, and how
is it used for the final classification?\footnotemark


\footnotetext{Although this work focuses on text classification, the findings in this section apply to any neural architecture which utilizes global max pooling, for both discrete and continuous domains. To our knowledge this is the first work that examines the assumption that max-pooling induces classifying behavior. Previously, \citet{DBLP:journals/corr/abs-1804-04438} showed that other assumptions to the functionality of max-pooling as deformation stabilizers (relevant only in continuous domains) do not necessarily hold true.}

\subsection{Informative vs. Uninformative Ngrams}
Consider the pooled vector $\m{p} \in R^m$ on which the classification is based. 
Each value $p_j = \text{ReLU}(\max_i \langle \m{u_i},\m{f_j} \rangle)$ stems from a
filter-ngram interaction, and can be traced back to the ngram $\m{u_i} = [\m{w}_i,...,\m{w}_{i+\ell-1}]$
that triggered it. Denote the set of ngrams contributing to $\m{p}$ as
$S_\m{p}$. Ngrams not in $S_{\m{p}}$ do not influence the decision of the
classifier. But what about the ngrams that are in $S_{\m{p}}$? 
Previous attempts in prediction-based interpretation of CNNs for text highlight
the ngrams in $S_\m{p}$ and their scores as means of explaining the
prediction. We take here a more refined view.
Note that the final classification does not observe the ngram identities
directly, but only through the scores assigned to them by the filters. Hence, the
information in $\m{p}$ must rely on the assigned scores.

Conceptually, we separate ngrams in $S_\m{p}$ into two classes,
\emph{deliberate} and \emph{accidental}.\\
\textbf{Deliberate} ngrams end up in
$S_{\m{p}}$ because they were scored high by their filter, likely because they
are \emph{informative} regarding the final decision. In contrast,
\textbf{accidental} ngrams end up in $S_\m{p}$ despite having a low score,
because no other ngram scored higher than them. These ngrams are likely
\textit{not informative} for the classification decision. Can we tease apart the
deliberate and accidental ngrams? We assume that there is
\emph{threshold} for each filter, where values above the threshold signal
informative information regarding the classification, while values below the
threshold are uninformative and can be ignored for the purpose of
classification. 
We thus search for the threshold that separate the two classes.
However, as we cannot measure directly which values $p_j$
influence the final decision, we opt instead for measuring \emph{correlation}
between $p_j$ values and the predicted label for the vector $\m{p}$.

The linearity of the decision function $\m{W}\m{p}$ allows to
measure exactly how much $p_j$ is weighted for the logit of label class
$k$. 
The class which filter $\m{f}_j$ contributes to is $c_j =
\operatorname*{arg\,max}_{k} W_{kj}$\footnotemark. We refer to class $c_j$ as the
\emph{class identity} of filter $f_j$.

\footnotetext{In the case of non-linear fully-connected layers, the question of how each feature contributes to each class is significantly harder to answer. Possible methods include saliency map methods or gradient-based methods. Recently, \citet{DBLP:journals/corr/abs-1805-08969} has attributed labels to filters using Bayesian inference and other image annotations.}

By assigning each filter a class identity $c_j$ and comparing it to the
predicted label we arrive at a \emph{correlation
label}---whether the filter's identity class matches the final decision by the
network. Concretely, we run the classifier over a set of texts, resulting in
pooled vectors $\m{p}^{i}$ and network predictions $c^i$. For each filter $j$ we
then consider the values $\m{p}_j^i$ and whether $c^i=c_j$. For each filter, we
obtain a dataset $(p_j^1, c^1=c_j),...,(p_j^D, c^D=c_j)$, and we look
for a threshold $t_j$ that separates $p_j^i$ for which $c^i=c_j$ from those
where $c^i\neq c_j$.
$$(X, Y)_j = \{(p_j^i, c^i=c_j) \mid j < m \ \& \ 
i < D\}$$

In an ideal case, the set is linearly separable and we can
easily separate informative from uninformative values: if $p_j^i > t_j$ then the
classifier's prediction agrees with the filter's label, and otherwise they
disagree. In practice, the set is not separable. We instead work with the \emph{purity} of
a filter-threshold combination, defined as the percentage of informative (correlative) ngrams which were scored above the threshold\footnotemark.  
Formally, given threshold dataset $(X, Y)$: \begin{multline*}
purity(f, t) = \\
\frac{\lvert\{(x,y) \in (X, Y)_f \ \mid \ x \geq t \ \& \ y = true \}\rvert}{\lvert\{(x,y) \in (X, Y)_f \ \mid \ x \geq t\}\rvert} 
\end{multline*}

\footnotetext{The purity metric can be considered as the precision metric for this task.}


We heuristically set the threshold of a filter to the lowest value that achieves
a sufficiently high purity (we experimentally find that a purity value of 0.75
works well).

\begin{figure*}[t]
\centering
\minipage{0.3\textwidth}
\centering
 (a) \includegraphics[width=\linewidth]{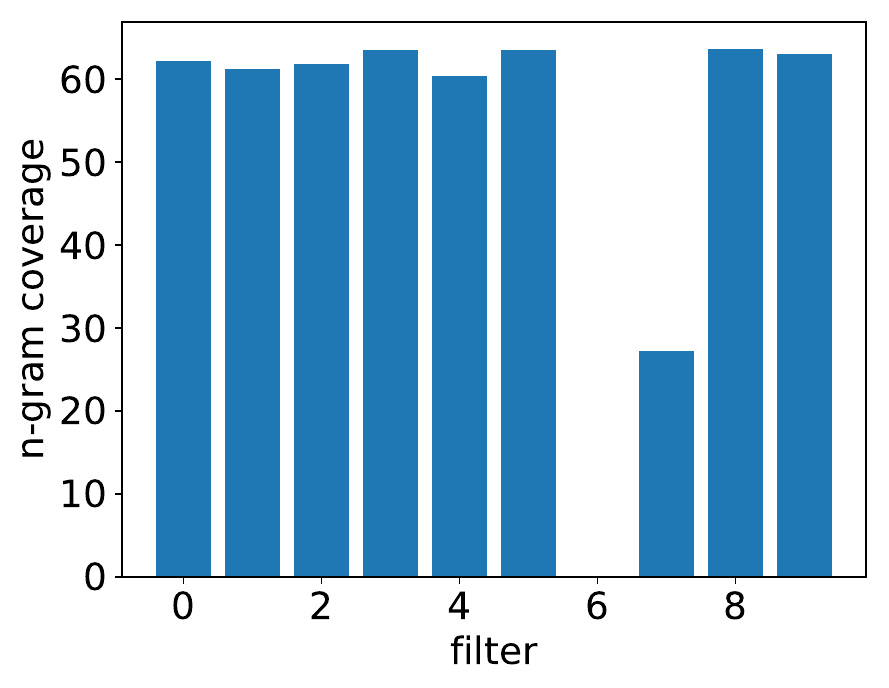}
\endminipage\hfill
\minipage{0.295\textwidth}
\centering
 (b) \includegraphics[width=\linewidth]{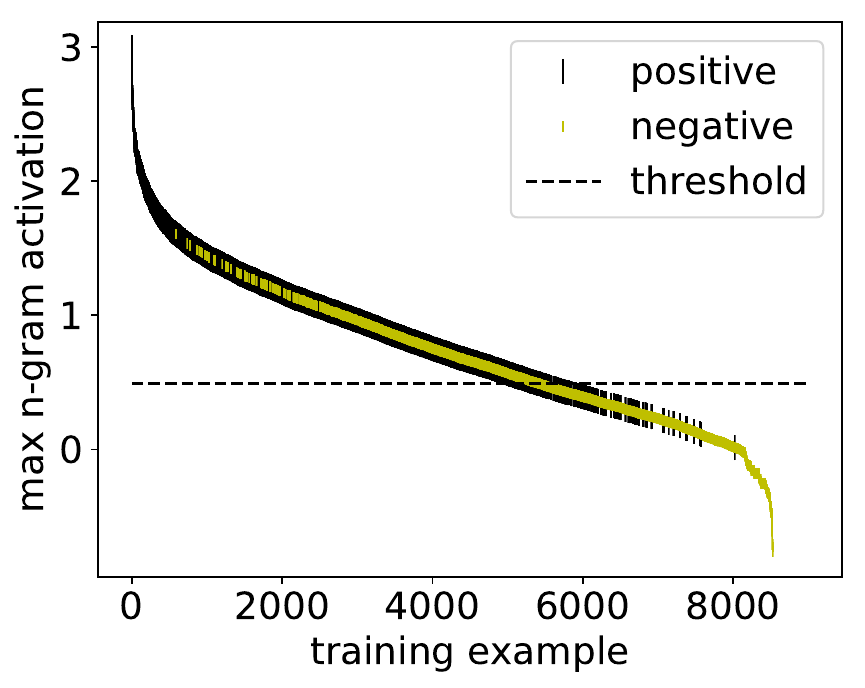}
\endminipage\hfill
\minipage{0.335\textwidth}%
\centering
 (c) \includegraphics[width=\linewidth]{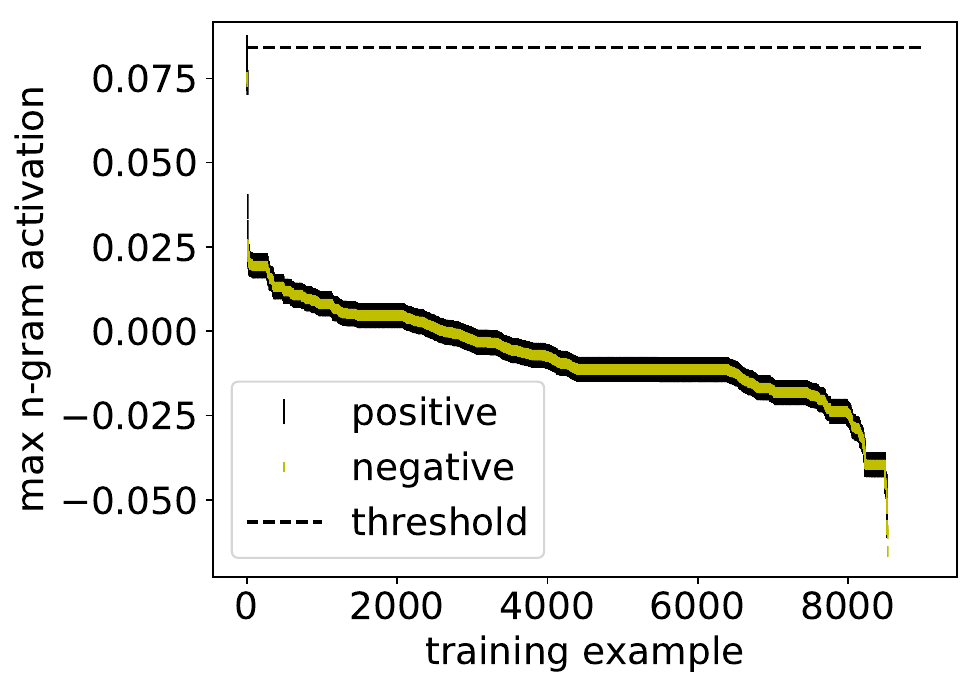}
\endminipage
    \caption{Visualization of informative and uninformative filters for the MR model and a universal purity of 0.75. In (a) we show the percentage of pooled ngrams which pass the threshold per filter. The threshold datasets of filters \#0 and \#6 are shown in (b) and (c) respectively.}
    \label{fig:threshold-bad-coverage}
\end{figure*}

In Figure \ref{fig:threshold-bad-coverage}b,c we show examples for threshold datasets for a model trained on the MR sentiment analysis task.


\paragraph{Threshold Effectiveness} 
\label{threshold-results}

We described a method for obtaining per-filter threshold values. But is the
threshold assumption---that items below a given threshold do not participate in
the decision---even correct?
To assess the quality of threshold obtained by our proposal and validate the
thresholding assumption, we discard values
that do not pass the threshold for each filter and observe the performance of
the model.  Practically, we replace the ReLU non-linearity with a threshold function: $$threshold(x, t)= 
\begin{cases}
x, & \text{if } x\geq t\\
0, & \text{otherwise}
\end{cases}$$

\begin{figure*}[ht]
\centering
\minipage{0.33\textwidth}
\centering
 (a) \includegraphics[width=\linewidth]{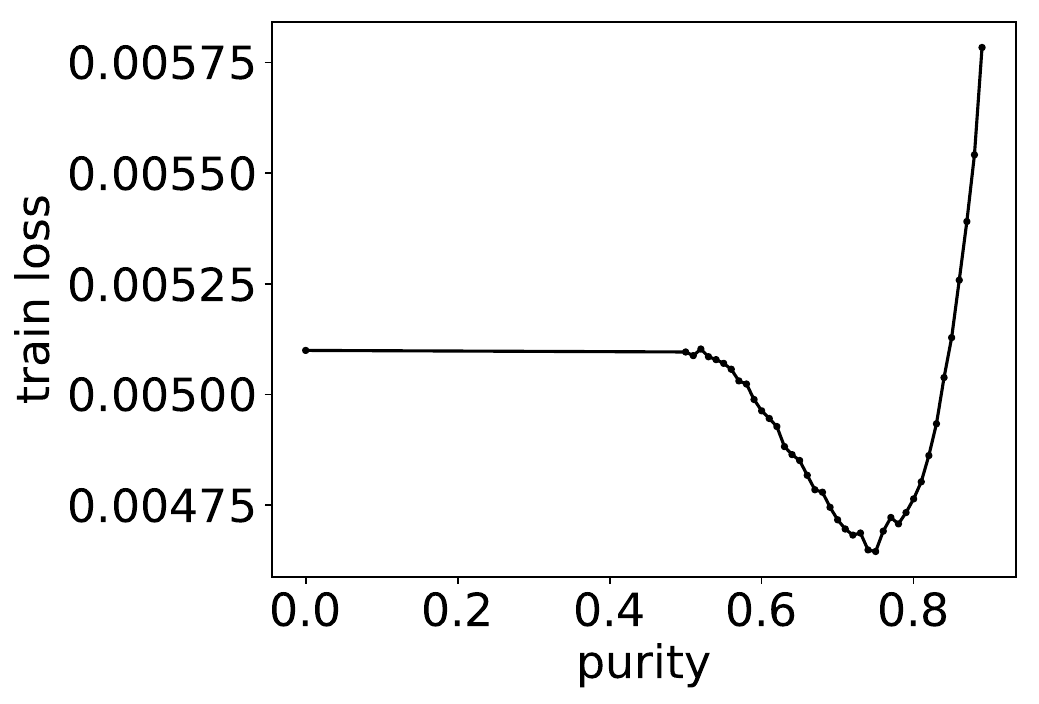}
\endminipage\hfill
\minipage{0.31\textwidth}
\centering
 (b) \includegraphics[width=\linewidth]{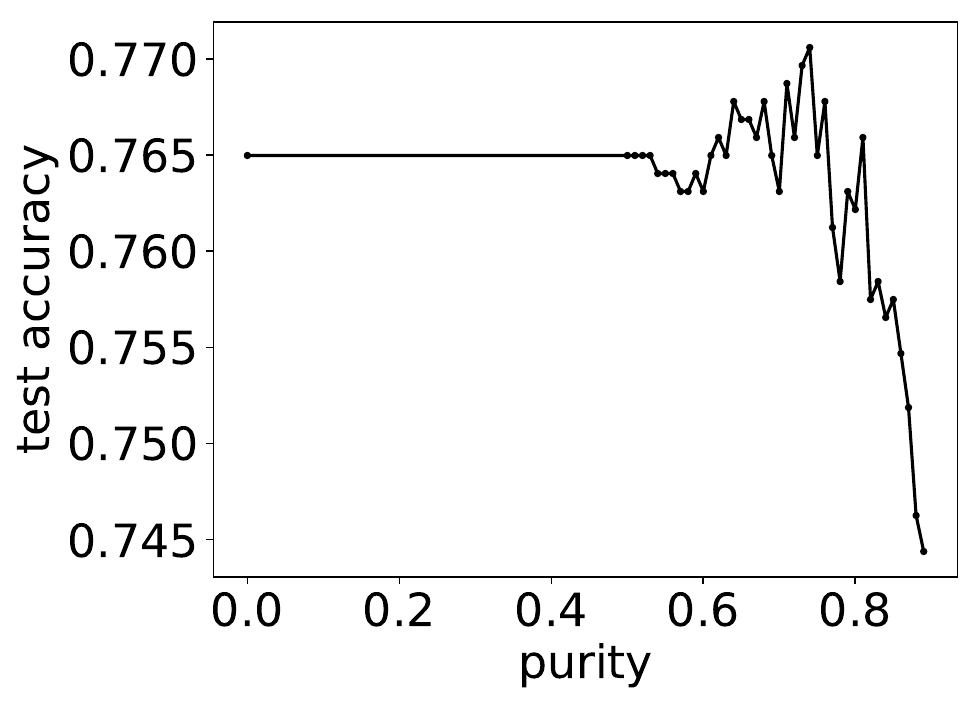}
\endminipage\hfill
\minipage{0.31\textwidth}%
\centering
 (c) \includegraphics[width=\linewidth]{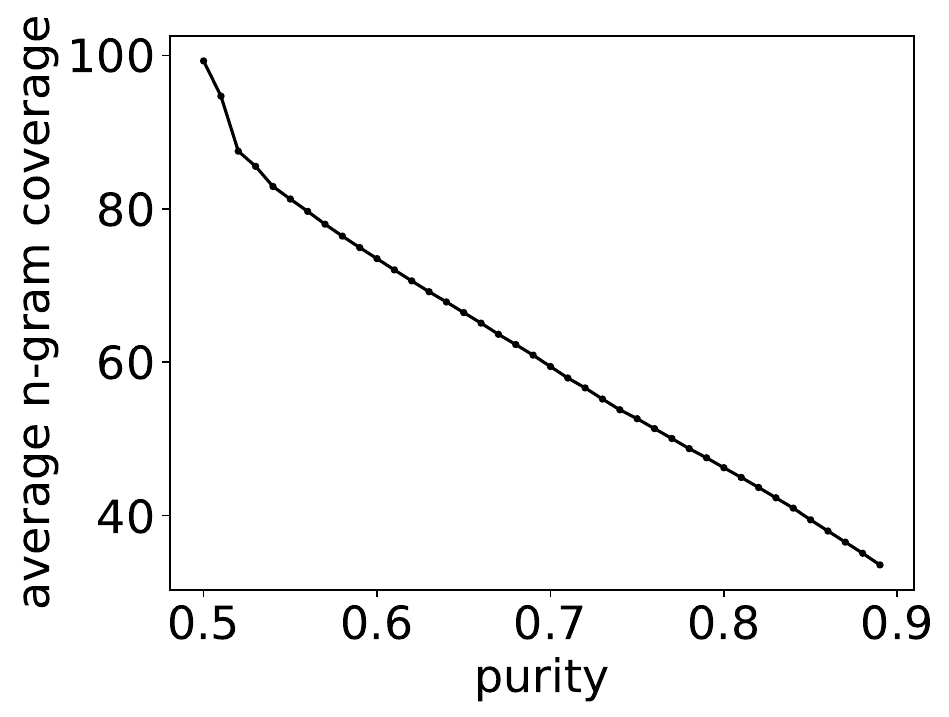}
\endminipage
    \caption{Evaluation results for identifying important ngrams on the MR model.}
    \label{fig:threshold-results}
\end{figure*}

Figure \ref{fig:threshold-results} presents the results on the MR dataset (we observed similar results on the Elec dataset).
where the threshold is set for each filter separately, based on a shared purity
value. If the thresholding assumption is correct and our way of deriving the
threshold is effective, we expect to not see a drop in
accuracy.  Indeed, for purity value of 0.75, we observe that the model performance
\emph{improves} slightly when replacing the ReLU with a per-filter threshold,
indicating that the thresholding model is indeed a good approximation for the
feature behavior. The percentage of informative (non-accidental) values in
$\m{p}$ is roughly a linear function of the purity (Figure
\ref{fig:threshold-results}c). With a purity value of 0.75\footnotemark, we discard roughly 44\% of
the values in $\m{p}$---and hence 44\% of the ngrams in $S_\m{p}$.

\footnotetext{We note that empirically and intuitively, the more filters we utilize in the
network, the less correlation there is between each filter's class and the final
classification, as the decision is being made by a greater consensus. This means
that demanding a higher purity will be accompanied by lower coverage, relative to other experiments, and more
ngrams will be discarded. The ``correct'' purity level for a filter then is a
function of the model and dataset used, and should be investigated using the
train or validation datasets.}

Not all filters behave in a similar way, however. In Figure
\ref{fig:threshold-bad-coverage} we show an example for a filter---\#6 in the
figure---which is especially uninformative: by applying the lowest threshold which satisfies a purity of 0.75, we discard 99.99\% of activations. Therefore in the experiments in Figure \ref{fig:threshold-results}, this filter is effectively unused, yet it does not cause loss in performance. In essence, the threshold classifier identified and effectively discarded a filter which is not useful to the model.

\paragraph{To summarize,} we validated our assumptions and shown empirically that global max-pooling indeed induces a functionality of separating important and not important activation signals using a latent (presumably soft) threshold. For the rest of this work we will assume a known threshold value for every filter in the model which we can use to identify important ngrams.

\section{What is captured by a filter?} \label{homogeneity}
Previous work looked at the top-k scoring
ngrams for each filter. However, focusing on the top-k does not tell a complete story. We insead look at
the set of deliberate ngrams: those that pass the filter's threshold value.
Common intuition suggests that each filter is
\emph{homogeneous} and specializes in detecting a specific classes of ngrams.
For example, a filter may specializing in
detecting ngrams such as ``had no issues'', ``had zero issues'', and ``had no
problems''. We challenge this view and show that filters often specialize in
multiple distinctly different semantic classes by utilizing activation patterns
which are not necessarily maximized. We also show
that filters may not only identify good ngrams, but may also actively supress
bad ones.

\subsection{Slot Activation Vectors}

As discussed in Section \ref{ff-refresh}, for each ngram $\m{u} =
[\m{w_1},...,\m{w_{\ell}}]$ and for each filter $\m{f}$ we calculate the score
$\langle \m{u},\m{f} \rangle$. The ngram score can be decomposed as a sum of
individual word scores by considering the inner products between every word
embedding $\m{w_i}$ in $\m{u}$ and every parallel slice in $\m{f}$:
$$ \langle \m{u},\m{f} \rangle = \sum_{i=0}^{\ell-1}{\langle \m{w_{i}}, \m{f_{id:i(d+1)}} \rangle}$$

We refer to slice $\m{f_{id:i(d+1)}} $ as
\textit{slot $i$} of the filter weights, denoted as $\m{f(i)}$. Instead of taking the sum of these
inner products, we can instead interpret them directly---saying that $\langle
\m{w_{i}}, \m{f(i)} \rangle$ captures how much slot $i$ in $\m{f}$ is
activated by the $i$th word in the ngram\footnotemark.

We can now move from examining the activation of an ngram-filter pair $\langle
\m{u}\coloneqq[\m{w_1};...;\m{w_\ell}], \m{f}\rangle$ to examining its \emph{slot activation
vector}: $(\langle \m{w_1},\m{f(1)}
\rangle,...,\langle\m{w_\ell},\m{f(\ell)} \rangle)$.
The slot activation vector captures how much each word in the ngram contributes
to its activation.

\footnotetext{
We note that this breakdown does not consider the filter's \textit{bias}, if one
is used. This bias is a single number (per filter) which is added to the sum of
slot activations to arrive at the ngram activation which is passed to the
max-pooling layer. Bias can be accommodated by appending an additional ``bias
word'' with an embedding vector of $[1,...,1]$ to every ngram. Regardless, as
this bias is identical for all ngrams for the filter in question, it has no role
in identifying which ngrams the filter is most similar to, and we can ignore it
in this context.
}

\subsection{Naturally occurring vs. possible ngrams}
We distinguish \emph{naturally occurring} or \emph{observed} ngrams, which are
ngrams that are observed in a large corpus, from \emph{possible} ngrams which
are any combination of $\ell$ words from the vocabulary. The possible ngrams are
a superset of the naturally occurring ones. Given a filter, we can find its
top-scoring naturally occurring ngram by searching over all ngrams in a
corpus. We can find its top-scoring possible ngram by maximizing each slot value
individually. We observe there is a big and consistent gap in scores between the
top-scoring natural ngrams and top-scoring possible ngrams. In our Elec model,
when averaging over all filters, the top naturally-occurring ngrams score 30\% 
less than the top possible ngrams. Interestingly, \emph{the top-scoring natural
ngrams almost never fully activate all slots in a filter}.

Table \ref{fig:topk2} shows the top-scoring naturally occurring and possible
ngrams for nine filters in the Elec model. In each of the top scoring natural
ngrams, at least one slot receives a low activation. Table \ref{fig:topk} zooms in on one of
the filters and shows its top-7 naturally occurring ngrams and top-7
most activated words in each slot. Here, most top-scoring ngrams maximize slot
\#3 with words such as \textit{invaluable} and \textit{perfect}, however some
ngrams such as ``\textit{works} as good'' and ``still \textit{holding} strong''
maximize slots \#1 and \#2 respectively, instead.

Additionally, most top-scoring words do not appear to be utilized in high-scoring ngrams at all.
This can be explained with the following: if a word such as \textit{crt} rarely or never appears in slot \#1 alongside other high-scoring words in other slots, then \textit{crt} can score highly with no consequence. Since an ngram containing \textit{crt} at slot \#1 will rarely pass the max-pooling layer, its score at that slot is essentially random. 



\begin{table*}[ht]
\centering
\resizebox{\linewidth}{!}{%
\begin{tabular}{c |l c c c c| l c | l c | l c | c} 
 & \multicolumn{5}{c|}{top ngrams} & \multicolumn{7}{c}{top words by slot}   \\
filter & ngram & score & \multicolumn{3}{c|}{slot scores} &
\multicolumn{2}{c}{slot \#1} & \multicolumn{2}{c}{slot \#2} &
\multicolumn{2}{c}{slot \#3} & sum \\ 
\hline
0 & poorly designed junk & 7.31 & 5.47 & 0.97 & 0.87 & poorly & 5.47 & displaying & 3.06 & landfill & 1.75 & 10.28 \\ 
1 & simply would not & 5.75 & 2.16 & 1.28 & 2.3 & chapters & 2.31 & avoid & 3.07 & impossible & 3.06 & 8.44 \\ 
2 & a minor drawback & 6.11 & 0.88 & 1.85 & 3.38 & workstation & 2.06 & high-quality & 3.82 & drawback & 3.39 & 9.27 \\ 
3 & still working perfect & 6.42 & 1.58 & 1.22 & 3.62 & saves & 2.52 & delight & 2.29 & invaluable & 4.19 & 9.0 \\ 
4 & absolutely gorgeous . & 5.36 & 1.09 & 3.84 & 0.42 & complain & 2.57 & gorgeous & 3.84 & expect & 1.22 & 7.63 \\ 
5 & one little hitch & 5.72 & 0.98 & 3.43 & 1.31 & path & 2.81 & delight & 4.09 & everyday & 2.64 & 9.54 \\ 
6 & utterly useless . & 6.33 & 2.03 & 3.49 & 0.81 & stopped & 2.77 & refund & 3.81 & disabled & 1.38 & 7.96 \\ 
7 & deserves four stars & 5.56 & 0.44 & 1.69 & 3.44 & excelente & 1.89 & crossover & 1.93 & incredible & 3.96 & 7.78 \\ 
8 & a mediocre product & 6.91 & 0.35 & 3.11 & 3.45 & began & 1.86 & mediocre & 3.11 & product & 3.45 & 8.42 \\ 
\end{tabular}}
\caption{Top ngrams and words by filter from a sample of nine filters from the Elec model. The average difference between the top natural ngram activation and the top possible ngram activation for this model is 2.5, or a 30\% average reduction.}
\label{fig:topk2}
\end{table*}

\begin{table*}[t]
\centering
\resizebox{\textwidth}{!}{%
\begin{tabular}{c | l c c c c|}
& \multicolumn{5}{c|}{top ngrams} \\
rank & ngram & score & \multicolumn{3}{c|}{slot scores} \\
\hline
1 & still working \textbf{perfect} & 6.42 & 1.58 & 1.22 & \textbf{3.62}  \\
2 & \textbf{works} - \textbf{perfect} & 5.78 & \textbf{1.91} & 0.25 & \textbf{3.62}  \\
3 & isolation proves \textbf{invaluable} & 5.61 & 0.39 & 1.03 & \textbf{4.19}  \\
4 & still near \textbf{perfect}  & 5.6 & 1.58 & 0.4 & \textbf{3.62} \\
5 & still working great & 5.45 & 1.58 & 1.22 & 2.65  \\
6 & \textbf{works} as good & 5.44 & \textbf{1.91} & 1.45 & 2.08  \\
7 & still \textbf{holding} strong & 5.37 & 1.58 & \textbf{1.81} & 1.98  \\
\end{tabular}
\begin{tabular}{l c | l c | l c}
\multicolumn{6}{c}{top words by slot} \\
\multicolumn{2}{c}{slot \#1} & \multicolumn{2}{c}{slot \#2} & \multicolumn{2}{c}{slot \#3} \\
\hline
saves & 2.52 	& delight & 2.29 	& \textbf{invaluable} & \textbf{4.19} \\
crt & 2.1 		& \textbf{holding} & \textbf{1.81} 	& \textbf{perfect} & \textbf{3.62} \\
beginner & 2.09 & welcome & 1.8 	& cm 	& 3.61 \\
mics & 2.08 	& dhcp & 1.72 		& pleasant & 3.38 \\
genius & 2.07 	& completely & 1.64 & simplicity & 3.14 \\
final & 2.01 	& cradle & 1.56 	& england & 3.09 \\
\textbf{works} & \textbf{1.91} & well-made & 1.51 & daily & 3.04 \\
\end{tabular}}
\caption{Top-k words by slot scores and top-k ngrams by filter scores from the Elec model. In bold are words from the top-k ngrams which appear in the top-k slot words - i.e. words which maximize their slot.}
\label{fig:topk}
\end{table*}


On naturally occurring ngrams, the filters do not achieve maximum values in all 
slots but only on some of them. Why?
We consider two hypotheses to explain this behavior: 
\begin{enumerate}[label=(\roman*)]
\setlength\itemsep{0.1em}
\item Each filter captures multiple semantic classes of ngrams, and each class 
has some dominating slots and some non-dominating slots (which we define as a \textit{slot activation pattern}). \label{heterogeneity-hypothesis}
\item A slot may not be maximized because it's not used to detect word
    existence, but rather lack of existence---ensuring that specific words do not occur. \label{negative-hypothesis}
\end{enumerate}
We investigate both hypotheses in Sections \ref{clustering} and
\ref{negative-ngrams} respectively.


\paragraph{Adversarial potential} We note in passing that this discrepancy in scores between naturally occurring
and possible ngrams can be used to derive adversarial examples that cause a
trained model to misclassify. By inserting a few seemingly random ngrams, we can
cause filters to activate beyond their expected range, potentially driving the
model to misclassification. We reserve this area of exploration for future work.

\subsection{Clustering (Hypothesis \ref{heterogeneity-hypothesis})} \label{clustering}


We explore hypothesis \ref{heterogeneity-hypothesis} by clustering
threshold-passing (naturally occurring) ngrams in each filter according to their
activation vectors.
We use Mean Shift Clustering
\cite{DBLP:journals/tit/FukunagaH75,DBLP:journals/pami/Cheng95}, an algorithm
that does not require specifying an a-priori number of clusters, and does not
make assumptions about their shapes. Mean Shift considers the feature
vectors as sampled from an underlying probability density function\footnote{Intuitively,
we can think of the sampling noise as the ngram embeddings, and the probability
distribution as defined by a function of the filter weights.}. 
%
%
Each cluster captures a different slot activation pattern. We use the cluster's
centroid as the prototypical slot activation for that cluster.

Table \ref{fig:clustering-example} shows a sample clustering output. The clustering algorithm identified two clusters:
one primarily containing ngrams of the pattern
\textit{DET INTENSITY-ADVERB POSITIVE-WORD}, while the second
contains ngrams that begin with phrases like \textit{go wrong}.\footnote{In the Yelp dataset,
\textit{go wrong} overwhelmingly occurs in a negated context such as
``can't go wrong'' and ``won't go wrong'', which explains why it is
detected by a positive filter.}

The centroids for these clusters capture the activation patterns well:
low-medium-high and high-high-low for clusters 1 and 2 respectively. 

\begin{table}[t]
\centering
\resizebox{\linewidth}{!}{%
\begin{tabular}{l c c c c}
ngram & slot \#1 & slot \#2 & slot \#3 & cluster \\
\hline
\hline
centroid & 0.75 & 1.97 & 2.79 & 1 \\
\hline
was super intriguing & 1.01 & 3.16 & 5.84 & 1 \\
am so grateful & 2.59 & 3.27 & 4.07 & 1 \\
overall very worth & 3.84 & 1.86 & 4.22 & 1 \\
also well worth & 1.83 & 3.06 & 4.22 & 1 \\
- super compassionate & 0.51 & 3.17 & 5.01 & 1 \\
a well oiled & 0.75 & 3.06 & 4.84 & 1 \\
\hline
\hline
centroid & 2.87 & 2.17 & 0.12 & 2 \\
\hline
go wrong bringing & 3.97 & 4.12 & 1.81 & 2 \\
go wrong pairing & 3.97 & 4.12  & 1.65 & 2 \\
go wrong when & 3.97 & 4.12  & -0.4 & 2 \\
\end{tabular}}
\caption{Example clustering results on the Yelp dataset. After applying thresholds, the ngrams for this filter were split into two clusters of sizes 83\% and 17\% respectively. The table shows top-scoring ngrams for this filter with their clustering results, sorted by their activation strength.}
\label{fig:clustering-example}
\end{table}


\paragraph{To summarize,} by discarding noisy ngrams which do not pass the filter's
threshold and then clustering those that remain according to their slot
activation patterns, we arrived at a clearer image of the semantic classes of
ngrams that a given filter specializes in capturing. In particular, we
reveal that filters are not necessarily homogeneous: a single filter may detect
several different semantic patterns, each one of them relying on a different slot
activation pattern.

\subsection{Negative Ngrams (Hypothesis \ref{negative-hypothesis})} \label{negative-ngrams}

Our second theory to explain the discrepancy between the activations of
naturally occurring and possible ngrams is that certain filter slots are not
used to detect a class of highly activating words, but rather to rule out a
class of highly negative words. We refer to these as \textit{negative ngrams}.

For example, Table \ref{fig:clustering-example} shows an ngram pattern for
which slot \#1 contains determiners and other ``filler'' tokens such as hyphens,
periods and commas with relatively weak slot activations. Hypothesis
\ref{negative-hypothesis} suggests that this slot may receive a strong
\textit{negative} score for words such as \textit{not} and \textit{n't},
causing such negated
patterns to drop below the threshold. Indeed, ngrams containing \textit{not} or
\textit{n't} in slot \#1 do not pass the threshold for this filter.

We are interested in a more systematic method of identifying these cases.
Identifying negative slot activations would be very useful for understanding
the semantics captured by a filter and the reasoning behind the dismissal of an ngram, as we discuss in Sections
\ref{model-interpretability} and \ref{predict-interpretability} respectively.

We achieve this by searching the below-threshold ngram space for ngrams
which are ``flipped versions'' of above-threshold ngrams. Concretely: Given
ngram $\m{u}$ which was scored highly by filter $\m{f}$, we search for low-scoring ngrams $\m{u'}$ such that the hamming distance between $\m{u}$ and $\m{u'}$ is low. By doing this for the top-k scoring ngrams per cluster, we arrive at a comprehensive set of negative ngrams. In Table \ref{fig:negative-example} we show a sample output of this algorithm.

Furthermore, we can divide negative ngrams into two cases: \begin{enumerate*}[label=\arabic*)]
\item Lowering the ngram score below the threshold by replacing high-scoring words with low-scoring words.
\item Lowering the ngram score below the threshold by replacing words with a
    low positive score with words with a highly-negative score.
\end{enumerate*} Case 2 is more interesting because it embodies cases where hypothesis \ref{negative-hypothesis} is relevant. Additionally, it highlights ngrams where a strongly positive word in one slot was negated with another strongly negative word in another slot. Table \ref{fig:negative-example} shows examples in bold.

In order to identify ``Case 2'' negative ngrams, we heuristically test whether the ``changed'' words' scores directly influence the status of the activation relative to the threshold: given an already identified negative ngram, if the ngram score---sans the bottom-k negative slot activations (considering a hamming distance of $k$ and given that there are $k$ negative slot activations)---passes the threshold, yet it does not pass the threshold by including the negative slot activations, then the ngram is considered a ``Case 2'' negative ngram.

\begin{table}[t]
\centering
\resizebox{\linewidth}{!}{%
\begin{tabular}{l c c c c}
ngram & slot \#1 & slot \#2 & slot \#3 & sum \\
\hline
\hline
'm really pleased & 2.59 & 1.86 & 5.05 & 9.5 \\
\hline
'm really not &  &  & -2.49 & 1.96 \\
'm really upset &  &  & -1.14 & 3.31 \\
\textbf{'m not pleased} &  & -3.4 &  & 4.24 \\
\hline
\hline
is extremely useful & 2.3 & 3.24 & 3.96 & 9.5 \\
\hline
\textbf{is extremely limited} &  &  & -2.8 & 2.74 \\
\textbf{is extremely noisy} &  &  & -2.77 & 2.8 \\
\textbf{is not useful} &  & -3.4 &  & 2.86 \\
\textbf{is only useful} &  & -2.82 &  & 3.44 \\
\hline
\hline
is surprisingly good & 2.3 & 4.32 & 2.8 & 9.42 \\
\hline
is not good &  & -3.4 &  & 1.7 \\
is only good &  & -2.82 &  & 2.28 \\
is no good &  & -1.88 &  & 3.22 \\
is probably good &  & -1.66 &  & 3.44 \\
\hline
\hline
am very satisfied & 2.01 & 2.17 & 5.09 & 9.26 \\
\hline
am very dissatisfied &  &  & -1.9 & 2.27 \\
am very disappointed &  &  & -1.87 & 2.3 \\
\textbf{am not satisfied} &  & -3.4 &  & 3.69 \\
\textbf{not very satisfied} & -2.6 &  &  & 4.66 \\
\end{tabular}}
\caption{Top-scoring ngrams from one filter from a model trained on the Elec dataset, and their accompanying lowest-scoring negative ngrams. We selected a hamming distance of 1 word. Bold ngrams are Case 2 negative ngrams.}
\label{fig:negative-example}
\end{table}

\section{Interpretability} \label{use-cases}
In this section we show two practical implications of the findings above:
improvements in both model-level and prediction-level interpretability of 1D
CNNs for text classification.

\subsection{Model Interpretability} \label{model-interpretability}
As in computer vision, we can now interpret a trained CNN model by
``visualizing'' its filters and interpreting the visible shapes---in other
words, defining a high-level description of what the filter detects.
We propose to associate each filter with the following items:
\begin{enumerate*}[label={\arabic*)}]
\item The class which this filter's strong signals contribute to (in the sentiment
    task: positive or negative);
\item The threshold value for the filter, together with its purity and coverages percentages (which essentially capture how informative this filter is);
\item A list of semantic patterns identified by this filter. Each list
    item corresponds to a slot-activations cluster. For each cluster we present the
    top-k ngrams activating it, and for each ngram we specify its total
    activation, its slot-activation vector, and its list of bottom-k negative
    ngrams with their activations and slot activations.
\end{enumerate*}
In particular, by clustering the activated ngrams according to their slot
activation patterns and showing the top-k in each clusters, we get a much
more refined coverage of the linguistic patterns that are captured by the filter.



\subsection{Prediction Interpretability} \label{predict-interpretability}
Previous prediction-based interpretation attempts traced back the ngrams from
the max-pooling layer. Here we improve these previous attempts by considering
only ngrams that pass the threshold for their filter. This results in a more concise and relevant explanation (Figure \ref{fig:threshold-results}). Figure \ref{fig:prediction-interpretability-eg} shows two examples. Note that in
example \#1, many negative-class filters were ``forced'' to choose an ngram in
max-pooling despite there not being strongly negative phrases---but those ngrams do not pass the threshold and are thus cleaned from the explanation.

Additionally we can use the individual slot activations to tease-apart the
contribution of each word in the ngram. Finally, we can also mark cases of negative-ngrams (Section \ref{negative-ngrams}), where an
ngram has high slot activations for some words, but these are negated by a
highly-negative slot and as a consequence are not selected by max-pooling, or are
selected but do not pass the filter's threshold.

\begin{figure}[t]
\centering
\resizebox{\linewidth}{!}{%
\begin{tabular}{$c^c^l^c^c^c}
\multicolumn{6}{p{1.2\linewidth}}{my \textbf{UNK fits perfectly} \textbf{. very well} made . nice looking and offers good protection} \\
\hline
\hline
filter & f-class & ngram & \multicolumn{3}{c}{slot scores} \\
\hline
0 & pos & PAD PAD my 			& 0.7  & 1.65 & 0.16 \\
\rowstyle{\bfseries}
1 & pos & . very well 			& 0.98 & 2.17 & 2.63 \\
2 & neg & PAD my UNK 			& 1.31 & -0.07 & 0.21 \\
3 & neg & UNK fits perfectly 	& 0.28 & 0.61 & 0.03 \\
4 & neg & looking and offers 	& 0.6  & 0.12 & 0.5  \\
5 & neg & good protection PAD 	& 0.52 & 1.6  & -0.01 \\
\rowstyle{\bfseries}
6 & pos & UNK fits perfectly 	& -0.06 & 2.36 & 1.82 \\
7 & neg & fits perfectly . 		& 1.34 & -0.71 & 1.47 \\
8 & neg & . very well 			& -0.01 & 1.97 & -0.55 \\
\rowstyle{\bfseries}
9 & pos & perfectly . very 		& 4.13 & 0.45 & -0.01 \\
\\
\multicolumn{6}{p{1.2\linewidth}}{this product \textbf{sucked was not} loud at all lights \textbf{did n't work} \textit{overall} \textbf{\textit{a bad} product} that 's UNK taking up space} \\
\hline
\hline
filter & f-class & ngram & \multicolumn{3}{c}{slot scores} \\
\hline
0 & pos & product sucked was 		& 0.12  & 2.05 & 0.1 \\
\rowstyle{\itshape}
1 & pos & overall a bad 			& 2.53 & 1.4 & -1.16 \\
2 & neg & lights did n't 			& -0.33 & 1.12 & 1.63 \\
3 & neg & PAD this product 			& -0.2 & 1.43 & 0.51 \\
\rowstyle{\bfseries}
4 & neg & did n't work 				& 1.21  & 0.97 & 2.65  \\
\rowstyle{\bfseries}
5 & neg & sucked was not 			& 0.98 & 0.59  & 1.32 \\
6 & pos & work overall a 			& -0.25 & 4.05 & -0.21 \\
7 & neg & was not loud 				& -0.33 & 2.85 & 0.52 \\
\rowstyle{\bfseries}
8 & neg & a bad product 			& -0.45 & 3.08 & 1.32 \\
9 & pos & PAD PAD this 				& 0.38 & 0.15 & 1.66 \\
\end{tabular}}
\caption{Examples predicted positive and negative respectively by a model trained on the Elec dataset, along with their explanations. Ngrams which passed the threshold are in bold, and case 2 negative ngrams are in italics. For clarity's sake we trained a small model which uses ten filters.}
\label{fig:prediction-interpretability-eg}
\end{figure}

\section{Conclusion} \label{discussion}

We have refined several common wisdom assumptions regarding the way in which CNNs
process and classify text. First, we have shown that max-pooling over time induces
a thresholding behavior on the convolution layer's output, essentially
separating between features that are relevant to the final classification and
features that are not. We used this information to identify which ngrams are
important to the classification. We also associate each filter with the class it
contributes to.
We decompose the ngram score into word-level scores by treating the
convolution of a filter as a sum of word-level convolutions, allowing us to
examine the word-level composition of the activation. Specifically, by
maximizing the word-level activations by iterating over the vocabulary, we
observed that filters do not maximize activations at the word-level, but instead
form slot activation patterns that give different types of ngrams similar
activation strengths. This provides empirical evidence that filters are not
homogeneous. By clustering high-scoring ngrams according to their
slot-activation patterns we can identify the groups of linguistic patterns
captured by a filter. We also show that filters sometimes opt to assign negative
values to certain word activations in order to cause the ngrams which contain them to
receive a low score despite having otherwise highly activating words.
Finally, we use these findings to suggest improvements to model-based and
prediction-based interpretability of CNNs for text.

\bibliography{emnlp2018}
\bibliographystyle{acl_natbib_nourl}

\end{document}